\documentclass{article}

\PassOptionsToPackage{numbers}{natbib}
\usepackage[preprint]{neurips_2022}

\usepackage{graphicx,wrapfig}
\usepackage{algorithmic}

\usepackage{comment}
\usepackage{amsmath}
\usepackage{amsthm}
\usepackage{soul}
\usepackage[ruled,noresetcount,vlined,linesnumbered]{algorithm2e}

\usepackage{todonotes}
\usepackage{xcolor}
\usepackage{setspace}
\usepackage{multirow}
\usepackage{bm,bbm}
\usepackage{paralist}
\usepackage{xr-hyper}

\makeatletter
\newcommand*{\addFileDependency}[1]{
  \typeout{(#1)}
  \@addtofilelist{#1}
  \IfFileExists{#1}{}{\typeout{No file #1.}}
}
\makeatother

\usepackage{enumitem}

 \newcommand{\cL}{\mathcal{L}}
\newcommand{\cM}{\mathcal{M}}

 \newcommand{\cY}{\mathcal{Y}}
 \newcommand{\cX}{\mathcal{X}}
 \newcommand{\RR}{\mathbb{R}}

\newcommand{\rev}[1]{{\color{purple}{#1}}}

\newcommand{\adv}{\text{adv}}

\DeclareMathOperator*{\argmax}{argmax}
\DeclareMathOperator*{\argmin}{argmin}

\usepackage[utf8]{inputenc} 
\usepackage[T1]{fontenc}    
\usepackage{hyperref}       
\usepackage{url}            
\usepackage{booktabs}       
\usepackage{amsfonts}       
\usepackage{nicefrac}       
\usepackage{microtype}      
\usepackage{xcolor}         
\usepackage{caption}
\usepackage{subcaption}

\title{Deadwooding: Robust Global Pruning for Deep Neural Networks}

\author{%
  Sawinder Kaur \\
  Department of EECS\\
  Syracuse University\\
  \texttt{sakaur@syr.edu} \\
  \And
  Ferdinando Fioretto \\
  Department of EECS\\
  Syracuse University\\
  \texttt{ffiorett@syr.edu} \\
  \And
  Asif Salekin \\
  Department of EECS\\
  Syracuse University\\
  \texttt{asalekin@syr.edu} \\
}

\begin{document}

\maketitle

\begin{abstract}
The ability of Deep Neural Networks to approximate highly complex functions is key to their success. This benefit, however, comes at the expense of a large model size, which challenges its deployment in resource-constrained environments. Pruning is an effective technique used to limit this issue, but often comes at the cost of reduced accuracy and adversarial robustness. This paper addresses these shortcomings and introduces \emph{Deadwooding}, a novel global pruning technique that exploits a Lagrangian Dual method to encourage model sparsity while retaining accuracy and ensuring robustness. The resulting model is shown to significantly outperform the state-of-the-art studies in measures of robustness and accuracy.
\end{abstract}

\section{Introduction} 

As deep learning models evolve and become more powerful, they also become larger and expensive to store and execute. This aspect challenges their deployment in resource constraint settings, including embedded systems and Internet of the Things, which require lightweight and real-time executable models.
To address this challenge, recent studies have developed a variety of methods to prune the relatively less important parameters of a neural network followed by a  \emph{fine-tuning} step that retrains the retained parameters to ensure high accuracy \citep{sehwag2019compact,aghli2021combining,zhang2018systematic}. 
The importance of training large, often over-parametrized models, followed by a pruning stage, has been emphasized in a wealth of studies \cite{zhu2017prune,han2015learning,li2016pruning} showing that the resulting pruned models perform overwhelmingly better, empirically, than dense models (of size comparable to the pruned models) trained from scratch.

While pruning can allow retaining high accuracy at a reduced model size, it has also been shown that the pruned models are often severely more susceptible to adversarial examples, when compared with the original models \cite{guo2019sparse} (see Appendix B for a detailed review). Since these size-reduced models are often deployed in security-critical contexts, including autonomous driving and home surveillance systems, poor model robustness can have disastrous consequences \cite{surveillance_adv,rossolini2022real}.  
{\em There is thus a critical need to corrobrate that pruned models ensure both accuracy and robustness to adversarial attacks.}

This paper addresses this need by introducing \textit{Deadwooding}, a novel robust global pruning paradigm which leverages Lagrangian duality to achieve the combined objective of model sparsity and robustness. 
The paper contributions are summarized as follows:
{\bf (1)} It casts the problem of robust model pruning to a constrained optimization problem that minimizes the empirical risk under a desired pruning degree and tolerance to adversarial attack. The resulting non-linear learning task is solved using Lagrangian dual learning method to compute an approximate solution that encourages the satisfaction of the desired pruning and robustness constraints.
{\bf (2)} While pruning, robustness is imparted using a novel proxy estimate that aims at increasing the distance between the projection of input samples and the decision boundary, thus, reducing their vulnerability towards adversarial attack. Notably, {\em this method is independent of any adversarial attack model}.
{\bf (3)} While global pruning retains accurate pruned models it may also generate disconnected networks at high pruning amounts \cite{blalock2020state}. To address this limitation, \emph{Deadwooding} uses a dedicated loss term to ensure network connectivity while globally selecting the parameters to be removed.
{\bf (4)} Next, the paper proposes a novel adaptation of knowledge distillation that incorporates adversarial training for the fine-tuning step. The resulting process retrains the non-pruned parameters to improve both accuracy and adversarial robustness.
{\bf (5)} To aid adversarial training during the fine-tuning step, the paper also introduces \textit{FGSM-looping}, a novel variation of the fast gradient sign method (FGSM) \cite{REN2020346}, which achieves comparable performance to the state-of-the-art (SoA) attack generation approaches with a significantly reduced computation time.
{\bf (6)} Finally, differently from the SoA robust pruning approaches that perform robust pruning only on already robust models, \emph{Deadwooding} produces robust pruned models irrespective of whether the original complex model has undergone adversarial training or not.

The extensive experimental evaluation on several settings and a benchmark dataset: CIFAR-10 illustrates significant improvements over state-of-the-art models. A detailed background description of this paper's relevant terms and concepts are provided in Appendix B.

\section{Related work and limitations}

The importance of model parameters towards model inference can be evaluated using several factors, including but not limited to the magnitude of their weights. The parameters with less magnitudes have been shown to have the least impact on the model's output \citep{Tang2015APB,magnitude_medium} and prediction accuracy. 
The majority of the SoA pruning studies \cite{han2015learning,sehwag2019compact,zhang2018systematic, han2015deep,Li2017PruningFF,savarese2020winning,Sanh0R20} (including the ones referred to in this paper) use magnitude-based parameter pruning.

While neural network model pruning has received much attention, 
the literature addressing the combined objective of pruning and robustness to adversarial examples is exceedingly more sparse. Notably, \citet{sehwag2019compact}(LWM) removed the model weights with the lowest magnitude iteratively while fine-tuning the model for adversarial robustness in each iteration, pruning a maximum of 90\% parameters.
\citet{zhang2018systematic} approached pruning using the alternating direction method of multipliers (ADMM), achieving a maximum of 71\% pruning. Recently, \citet{sehwag2020hydra} presented Hydra, which used an adversarial training objective during the pruning training step and fine-tuned the resulting pruned model to ensure high accuracy, achieving a maximum of 99\% pruning. 

\begin{wrapfigure}[11]{r}{8.5cm}
\vspace{-12pt}
  \centering
  \includegraphics[width=8.5cm]{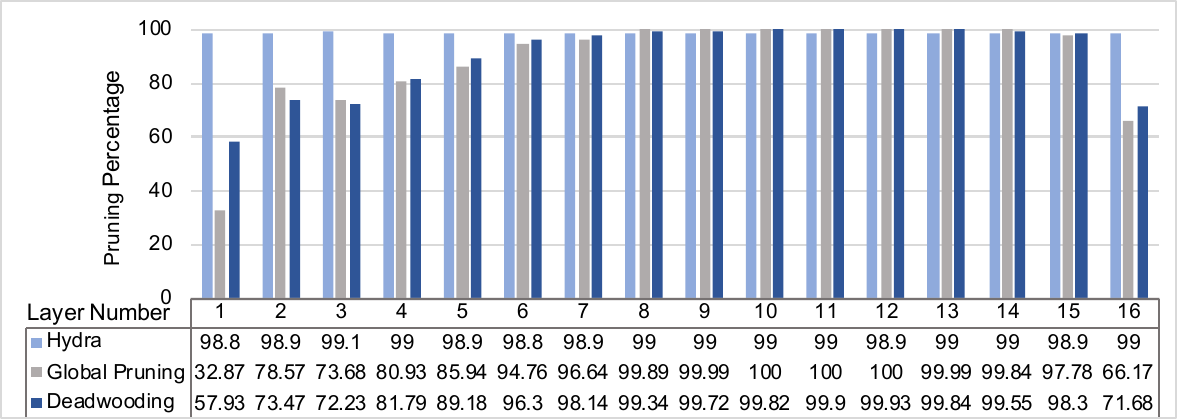}
  \caption{Per-layer pruning amounts for different approaches at $\sim$99\% pruning of VGG-16 for CIFAR-10 dataset}
  \label{fig:layer-wise}
\end{wrapfigure}



Although these works 
present important contributions to the state of the art, they all have the following two limitations: 
{\bf (1)} To avoid creating disconnected sub-networks, these methods all perform layer-wise pruning, which restricts the pruning amount for each layer to a fixed percentage. As we will show next, this strategy induces suboptimalities to the model performance.  
{\bf (2)} Existing robust pruning approaches assume that the original, complex model to be pruned, is already an adversarially robust model, and attempt at preserving such property.

Limitation {\bf (1)} is expressed in Figure \ref{fig:layer-wise}, which reports the amount of parameters removed, in each layer of VGG-16 at 99\% pruning for three approaches: {\bf (i)} a \emph{generic} least-weight magnitude based global pruning method (described in more details in Appendix B), {\bf (ii)} Hydra by \citet{sehwag2020hydra}, which performs layer-wise pruning, and {\bf (iii)} \emph{Deadwooding}. 
Notably, in both global pruning approaches (\emph{Deadwooding} and {\em generic}), the initial feature extraction layers and the final decision-making layers undergo less pruning than the middle layers. The generic global pruning removes parameters based on their magnitudes and the importance of a parameter is considered to be directly proportional to its magnitude \citep{han2015learning}. Thus, it indicates initial and final layer's parameters are more critical for accurate classifications. But, generic global pruning removes the layers 10--12 completely, resulting in a disconnected network. While Hydra \cite{sehwag2020hydra} generates a connected pruned network by removing an equal amount of parameters from each layer, it also removes more parameters in the 
critical layers than \emph{Deadwooding}; thus, \emph{Deadwooding} generates pruned models having higher accuracy and robustness.

\section{Problem settings and model pruning goals}
\label{sec:settings_and_goals}

The paper considers datasets $D$ of $n$ individual data points $(X_i, Y_i)$, with $i \!\in\! [n]$ drawn from an unknown distribution. Therein, $X_i \!\in\! \mathcal{X}$ is a feature vector and $Y_i \!\in\! \mathcal{Y}$ is a label. 
Given a classifier $\cM_\theta \!:\! \cX \!\to\! \cY$, where $\theta \in \mathbb{R}^k$
is a real valued vector describing the model parameters, the goal is 
to learn a reduced classifier $\cM_{\theta^\downarrow} \!:\! \cX \!\to\! \cY$ 
with $\theta^\downarrow \in \RR^{k'}$ and $k' \ll k$, which is 
substantially smaller than the original classifier while retaining 
\emph{high benign accuracy} and ensuring \emph{robustness}.
The model quality is measured in terms of a non-negative \emph{loss function}
$\ell : \cY \times \cY \to \mathbb{R}_+$, which, in this work is the \emph{cross-entropy loss} \citep{cel}. The problem is that of 
minimizing the empirical risk function:
\begin{equation}
\label{eq:ERM}
\min_{\theta} {\cal L}(\theta; D) =
\frac{1}{n} \sum_{i \in [n]} \ell\left(\cM_\theta(X_i), Y_i \right).
\end{equation}
The resulting model performance over the test set is referred to as \emph{benign accuracy}.

\emph{Adversarial robustness} refers to a model’s ability to resist to adversarial examples\cite{szegedy2013intriguing}: inputs that are produced by 
perturbing them slightly with a perturbation budget ($\epsilon\! >\! 0$) in a way that causes them to be misclassified by the model. The prediction accuracy of the model against adversarial examples \cite{chakraborty2018adversarial,akhtar2018threat,madry2018towards}  refers to its adversarial robustness, and it is measured in terms of an adversarial loss function, as defined by \citet{sehwag2020hydra}:
\begin{equation}
    \cL_{\adv}(\epsilon, \theta; D) = 
    \nicefrac{1}{n} \sum_{i \in [n]} \ell_{\adv}(\cM_{\theta}(X_i \oplus \epsilon)
    , Y_i).
    \label{adv-training-background}
\end{equation}
Therein, $\oplus$ is an adversarial attack generation method and $\ell_\adv$ is an adversarial loss assessing the effect of the adversarial attack. 


\textbf{Problem statement.} Given a pruning target $k'$, the desired goal is to learn a reduced model $\cM_{\theta^\downarrow}$ with $\theta^\downarrow \subset \theta$ that has high benign accuracy while being robust. This ideal objective is formalized as:
\begin{subequations}
\label{eq:p_prune}
\begin{align}
    \label{eq:5a}
\textstyle  {\bm{b}}^{\downarrow} = \argmin_{\bm{b}}\;\; &  
    {\cal L}(\theta\,\bm{b}; D) \\
    \label{eq:5b}
    \text{such that:}\;\;&  \cL_{\adv}(\epsilon, \theta\,\bm{b}; D) = 0\\
    \label{eq:5c}
                &   \| \bm{b} \|_2 = k'\\ 
    \label{eq:5d}
                &   \bm{b} \in \{0,1\}^k,
\end{align}
\end{subequations}
where $\bm{b}$ is a $k$-dimensional binary vector. 
In the above, the constraint \eqref{eq:5b} imposes the pruned model to be robust to adversarial attacks and \eqref{eq:5c} defines a \emph{capacity constraint} used to construct the pruned model. The optimal pruned parameters $\theta^\downarrow$ can be retrieved as $\theta\bm{b}$.

\section{Robust pruning: a Lagrangian dual view}\label{section:lagrangian-dual}
Notice that the ideal problem above is an NP-hard Mixed Integer Non-Linear program. Finding a reduced model $\cM_{\theta^\downarrow}$ through Program \eqref{eq:p_prune} is thus not computationally viable, therefore, this section specifies a \emph{Lagrangian relaxation} of the program above and proposes a Lagrangian Dual approach to enforce the problem constraints during training. 

In \emph{Lagrangian relaxation}, the constraints of an optimization problem are relaxed into the objective function using \emph{Lagrangian multipliers} to capture the penalty induced by violating them \cite{lagrangian_ff}. 
The Lagrangian function of Problem \eqref{eq:p_prune} is:
\begin{equation}
\label{eq:lagrangian_relaxation}
{\cal L}_\lambda(\bm{b}) = {\cal L}(\bm{b}) + \lambda_a |\cL_{\adv}(\bm{b})| 
+ \lambda_p | \|\bm{b}\|_2 - k'|
\end{equation}
where parameters $\theta$ and $D$ are omitted for clarity of explanation. Therein, variables $\bm{b}$ have been relaxed in $[0,1]\subseteq \RR$ and the terms $\lambda_a$ and $\lambda_p$
describe the Lagrangian multipliers associated, respectively, with the adversarial loss and the pruned model capacity constraints. 

For multipliers $\bm{\lambda} = (\lambda_a, \lambda_p)$ solving the optimization problem above produces an approximation $\cM_{{\theta}^\downarrow(\lambda)}$ of $\cM_{\theta^\downarrow}$. The Lagrangian dual computes the optimal multipliers, i.e.,:
\begin{equation}
    \bm{\lambda}^* = \argmax_{\bm{\lambda}} \min_{\bm{b}} {\cal L}_\lambda(\bm{b})
\end{equation}
to obtain $\tilde{\cM}^* \!\!=\! \cM_{{\theta}^\downarrow(\lambda^*)}$, the strongest Lagrangian relaxation of Problem \eqref{eq:p_prune}. 
Learning $\tilde{\cM}^*$ relies on an iterative scheme that
interleaves the learning of a number of Lagrangian relaxations (for
various multipliers) with a subgradient method to learn the best
multipliers:
\begin{subequations}
\label{eq:dual}
\begin{align}
    \bm{b}^{t+1} &= \argmin_{\bm{b}} 
    {\cal L}_{\lambda^t}(\bm{b}\theta; D) \label{eq:L1} \\
    \bm{\lambda}^{t+1} &= \big(
    \lambda^t_a \!+\! \rho\, | {\cal L}_\adv(\theta\bm{b}^{t+1}; D)|,\lambda^t_p \!+\! \rho\, \left| \| \bm{b}^{t+1} \|_2 \!-\! k' \right| 
    \big), \label{eq:L2} 
\end{align}
\end{subequations}
\noindent
where $\rho > 0$ is the Lagrangian step size. In the implementation, step \eqref{eq:L1} is approximated using a gradient-based method.

While a Lagrangian relaxation of Problem \eqref{eq:p_prune} does not obviously ensure finding the minimizer which satisfies the problem constraints exactly, as we will show in Section \ref{section:experiments}, the Lagrangian dual method exploited in this paper is highly effective in building robust pruned models. The Lagrangian dual learning method to prune a model is a novel contribution of this work.

\section{Deadwooding}
\emph{Deadwooding} consists of two phases: {\bf (1)} Model Pruning and 
{\bf (2)} Fine-tuning. First, the Lagrangian Dual based pruning approach forces the least useful $\frac{k-k'}{k}\%$ (i.e., desired amount of pruning) network parameters to approach zero while ensuring high accuracy and robustness. These parameters are then removed with minimum impact on the 
model accuracy (see Section \ref{sec:Model_pruning}). Next, an adaptation of the knowledge distillation is used as a fine-tuning step (see Section \ref{sec:fine_tuning}).

\subsection{Model pruning step}
\label{sec:Model_pruning}

\begin{wrapfigure}{r}{8 cm}
\vspace{-40pt}
\begin{flushright}
\begin{minipage}{7.8cm}
\begin{algorithm}[H]
{\small
  \caption{Robust Pruning}
  \label{alg:learning}
  \setcounter{AlgoLine}{0}
  \SetKwInOut{Input}{input}

  \Input{$M_\theta$: Pre-trained model
  $~\diamond~ k'$: Target pruning amount
  $~\diamond~ D=(X_i, Y_i)_{i=1}^n$: Training data
     $~\diamond~  \alpha, \rho=(\rho_0, \rho_1,\ldots)$: Optimizer and Lagrangian step sizes
       $~\diamond~ W$: Parameter weight threshold.
       \!\!\!\!\!\!\!\!\!\!\!\!\!\!\!\!\!\!\!\!\!\!}
  \label{line:1}
  $\lambda_a^0 = \lambda_p^0 \gets 0 $\\
  \While{$\|\bm{b}\|_{\geq W} \leq k'$ [epoch $t = 0, 1, \ldots$]}{ 
   
  \label{line:2}
      \ForEach{$(X, Y) \in D$} { 
      \label{line:3}
      $\hat{Y} \gets {\cal M}_{\bm{b}\theta (\lambda^t)}(X)$\\
      \label{line:4}
      $\bm{b} \gets \bm{b} - \alpha \nabla_{\bm{b}} \left(
            \ell(\hat{Y}, Y) + \lambda^t_a | \ell_\adv(X ) |
                             + \lambda^t_p | \|\bm{b}\|_2 - k' | \right)$
      \label{line:5}
    }
    $\lambda^{t+1}_a \gets \lambda^t_a + \rho_t | {\cal L}_\adv(\bm{b}\theta, D) |$\\
    \label{line:6}
    $\lambda^{t+1}_p \gets \lambda^t_p + \rho_t | \|\bm{b}\|_{\geq W} - k' |$
    \label{line:7}
  }
}
\end{algorithm}
\end{minipage}
\vspace{-12pt}
\end{flushright}
\end{wrapfigure}

The pruning step implements the robust pruning objective through the Lagrangian dual method described in Equation \eqref{eq:dual}. The process is described in Algorithm \ref{alg:learning}. 
Given the input dataset $D$, the optimizer step size $\alpha > 0$, 
and a Lagrangian step size $\rho_t$, the Lagrangian multipliers are 
initialized in line \ref{line:1}. The training is performed until the required number of parameters are below a pre-decided threshold $W$ (line \ref{line:2}), and each epoch $t$ optimizes the reduced model 
parameters ${\theta}^\downarrow = \bm{b} \theta$ of the classifier
 ${\cal M}_{\bm{b}\theta (\lambda^t)}$ using the Lagrangian multipliers
  $\lambda^t$ associated with current epoch (lines \ref{line:3}--\ref{line:5}). 
Finally, after each epoch, the Lagrangian multipliers are updated according 
to a \emph{dual ascent} rule \cite{boyd2011distributed} (lines \ref{line:6} and \ref{line:7}).
The resulting model statisfies the required capacity constraints (given the imposed threshold factor). This training step is thus followed by the removal of the remaining $k-k'$ parameters with corresponding $\bm{b}$ values smaller than $W$. 

The following subsections describe the three components of the pruning loss Equation \eqref{eq:lagrangian_relaxation} (Algorithm \ref{alg:learning} line \ref{line:5}) and their relation with achieving robustness and pruning.

\noindent\textbf{Adversarial robustness proxy.} 
To incorporate robustness during the pruning step, Deadwooding uses an adversarial proxy loss function that aims at maximizing the distance of the projection of the input samples to their decision boundary. 
Intuitively, samples farther away from the decision boundary are less 
sensitive to input perturbations under the lens of the model output. 
Following \citet{tran2021differentially}, we approximate the distance of an input sample $X$ to the decision boundary for a $K$-class classifier 
using the output logit values: 
\begin{equation}
\textstyle
   \ell_{\adv}(X) = 1 - \sum_{k=1}^K {\cM}_{\theta,k}^2 (X).
   \label{Dist-eq}
\end{equation}
The term $\cM_{\theta,k}(X) \in [0,1]$ denotes the $k^{th}$ logit value associated  
with input $X$ given the model with parameters $\theta$. The above rely on the fact that, samples close to the decision boundary are associated with uncertain model decisions. Thus, the logit values of 
samples that are close to the decision boundary tend to be similar and approach $1/K$, as the samples get closer to the decision boundary. This leads to a high ${\ell}_\adv$ value.
Conversely, samples which are farther away from the decision boundary, are predicted with higher confidence and, thus, one of the logit values will be much higher than the others, leading to a 
low $\ell_{\adv}$ value. Thus, minimizing $\cL_{\adv}$ leads to learning models which separates the input samples from their decision boundary as much as possible.

\noindent\textbf{Pruning constraint proxy.} 
Note that the pruning constraint $\|\bm{b}\|_2 - k'$ introduced in the 
Lagrangian loss function effectively models a \emph{Ridge} regularization. The regularization term enforces sparsity in the pruned network.
The Lagrangian dual model enforces this regularization term with 
strength $\lambda_p$ which is proportional to the constraint violation 
(difference to the desired pruning amount).

\noindent\textbf{Global pruning.}\label{approach:globalpruning}
As mentioned earlier, existing robust pruning studies are limited to layer-wise pruning, since global pruning may generate disjoint subnetworks in high pruning regimes (see Figure \ref{fig:layer-wise}). A simple, yet effective way to address this limitation is by adopting the loss term $\cL$ (which measures the prediction accuracy) in Equaion \eqref{eq:lagrangian_relaxation}. In the absence of an active path, the pruned model’s loss ${\cal L}_\lambda$ will increase significantly. 
During the model pruning step, the pruned network maintains an active path through the retained parameters ${\theta}^\downarrow = \bm{b} \theta$ to minimize ${\cal L}_\lambda$, thus enabling global pruning while also ensuring a connected pruned network.

\subsection{Fine-tuning step}
\label{sec:fine_tuning}

Conventionally, pruning is followed by a retraining step over the non-pruned parameters (known as fine-tuning) that helps to relearn the information lost by removal of the pruned parameters. Interestingly, \citet{KD_prune} showed that a pruned model fine-tuned using knowledge distillation (KD) exhibits better prediction accuracy than one subject to natural re-training. 
Vanilla KD \cite{KD_main} aims to transfer the knowledge learned by one (teacher) model, to another (student) model. The student is trained to achieve high accuracy mimicking the teacher model by learning response-based knowledge. 

The KD loss function $\cL_{\rm KD}$ has two components: a hard loss, $\cL$, and a soft loss, $\cL_{\rm soft}$. 
The former accounts for student model's prediction errors whereas the latter enables the student to learn the relation among different classes through a softmax computed at high temperature values (see Appendix B for additional details). 
In its fine-tuning step, \emph{Deadwooding} uses a modified KD step, which includes an additional component to account for adversarial loss $\cL_{\adv}(\epsilon, \theta; D)$ defined as in Equation \eqref{adv-training-background}. The original, complex, model is referred to as the teacher and the pruned model as the student. The overall fine-tuning loss function is defined as:
\begin{equation}
    \cL_{\text{fine-tune}}(\epsilon) = \alpha \cL + \beta \cL_{\text{soft}} + \gamma \cL_{\adv}(\epsilon),
    \label{loss-kd}
\end{equation}
where $\theta$ and $D$ have been removed for clarity of explanation and 
$\alpha, \beta, \gamma$ are trainable hyper-parameters. In the experiments, these values were obtained using the hyper-parameter tuning tool Optuna \cite{akiba2019optuna}. 
It is important to note here that Lagrangian-dual approximation is applicable to only constrained optimization problems and thus, not directly applied to optimize the coefficients of Equation \eqref{loss-kd}.

The fine tuning step requires the generation of adversarial examples to evaluate the loss component $\cL_\adv$.
To do so, \emph{Deadwooding} uses a simple adaptation of FGSM, named FGSM-looping, that instead of using a single perturbation value $\epsilon$ to generate the adversarial examples, it iterates through a set of values $E = (\epsilon_1, \epsilon_2, \ldots)$ with each $\epsilon_i \in (0, \epsilon_{\max}]$
with $\epsilon_{\max}$ being the maximum perturbation allowed. 
This helps to generate multiple perturbed samples around a point within an $L_\infty$-ball of radius $\epsilon_{\max}$, whereas vanilla FGSM generates adversarial samples at the boundary of the region of interest. 
The evaluation illustrates that using FGSM-looping during fine-tuning reduces computation time while achieving {\sl eba} and {\sl era} comparable to SoA approaches. Further details are provided in Appendix D.3.

\section{Experiments}
\label{section:experiments}

This section compares the \emph{Deadwooding}'s performance, in terms of accuracy and adversarial robustness, against the current SoA pruning method Hydra \cite{sehwag2020hydra}. 
Notice that Hydra is the only SoA robust pruning baseline available and, thus, the most relevant to our analysis. 

\textbf{Settings and evaluation metrics.}~ 
The evaluations uses the benchmark dataset: CIFAR-10 and \textit{VGG-16} network architecture. 
Further detail on experimental setup, system, and configurations are discussed in Appendix A.

This paper uses two metrics to measure the performance of pruned networks: \emph{Empirical Benign Accuracy ({\sl eba})} and \emph{Empirical robustness accuracy ({\sl era})}, which are defined by the SoA baseline \cite{sehwag2020hydra}. {\sl eba} is the percentage of correctly classified benign (i.e., not adversarially perturbed) examples, where {\sl era} is the percentage of correctly classified adversarial examples generated with an adversarial attack $\oplus$.
Following the observation by \citet{carlini2019evaluating}, suggesting that using the same $\oplus$ to attain and evaluate robustness of a network can lead to biased results, \emph{Deadwooding} uses FGSM-looping as $\oplus$ in Equation \eqref{adv-training-background} to impart robustness during fine-tuning step, and PGD attack \cite{REN2020346} as $\oplus$ to evaluate the pruned model’s robustness (i.e., {\sl era}). However, the paper also evaluates the pruned models’ robustness with several different (and stronger) attacks such as AutoAttack \cite{auto-attack}, FAB \cite{fab-attack} and DeepFool \cite{deepfool-attack} as $\oplus$. These extended results are reported in Appendix E.2.

\subsection{Robust network pruning evaluation}



This section evaluates the ability of \emph{Deadwooding} to retain accuracy and ensure robustness. Table \ref{main_table} compares \emph{Deadwooding} with Hydra when pruning \emph{robust} original models. Table \ref{main_table} reports mean values for 10 different repetitions and for three pruning constraints: $90$, $95$, and $99$, indicating the percentage of parameters to be pruned from the original networks (PT(=$0$\%)).

\begin{table}[t!]
    \centering
    \begin{tabular}{p{1.6cm} |p{1.2cm} | rr rr rr}
    \toprule
    {\textcolor{violet}{Datasets}}
    &{Pruning} 
    & \multicolumn{2}{c}{Hydra} 
    & \multicolumn{2}{c}{Ours }
    & \multicolumn{2}{c}{$\Delta$}\\
    
    \& \textcolor{blue}{Model} &\%age & \multicolumn{2}{c}{ ($\cM_{\theta^\downarrow}^{r(H)}$)} 
    & \multicolumn{2}{c}{ ($\cM_{\theta^\downarrow}^r$)}
    & \multicolumn{2}{c}{ ($\cM_{\theta^\downarrow}^r$,$\cM_{\theta^\downarrow}^{r(H)}$)}\\
    && {\sl eba} & \rev{{\sl era}} & {\sl eba} & \rev{{\sl era}}& {\sl eba} & \rev{{\sl era}}\\
    \midrule
    &PT(=0\%)& & & 82.7 & \rev{51.9} &\\           \textcolor{violet}{CIFAR10} &90\% & 80.5&\rev{49.5} &\textbf{83.56}&\rev{\textbf{59.68}} &+3.06&\rev{+10.18}\\
    \textcolor{blue}{VGG16} &95\% & 78.9&\rev{48.7} & \textbf{82.87}&\rev{\textbf{59.8}}  &+3.97&\rev{+11.1}\\
    &99\% & 73.2&\rev{41.7}& \textbf{80.87}&\rev{\textbf{53.1}} &+7.67&\rev{+11.4}\\

    \bottomrule
\end{tabular}
\caption{Comparison of pruned models obtained using Hydra and \emph{Deadwooding} starting from robust original model ($M_\theta^r$) VGG-16  trained for CIFAR-10 based on {\sl eba} and {\sl era}. $\Delta$ represents the performance gain by \emph{Deadwooding}.}
\label{main_table}
\end{table}

Therein, $\cM_{\theta^\downarrow}^{r}$, and $\cM_{\theta^\downarrow}^{r(H)}$ denote  the pruned models resulting from \emph{Deadwooding} and Hydra starting $\cM_\theta^r$, respectively. Key observations from Table \ref{main_table} are:

\par
{$\bullet$} \emph{Deadwooding extracts pruned models that are more robust than their original complex counterparts}. 
\emph{Deadwooding} achieves a mean improvement in by {\sl era} $5.63$ ($\Delta(\cM_{\theta}^r$,$\cM_{\theta\downarrow}^r)$) percentage points with respect to the original models (PT=$0$\%).
\par
{$\bullet$} \emph{Deadwooding produces highly accurate pruned models}. 
While it is folklore that pruning, in general, reduces overfitting and, thus, may result in accuracy improvement \cite{bartoldson2020generalizationstability}, extreme pruning may generate networks with insufficient parameters to model the classification function, resulting in a loss of {\sl eba}. 
Surprisingly, the reported results show that \emph{Deadwooding} achieves pruned models having {\sl eba} comparable to original model. {\em This is significant, as large decreases in {\sl eba} may limit the usability of the pruned models, especially in applications such as health monitoring and autonomous systems.} 
\par

\textbf{Deadwooding's comparison with SoA.} A detailed comparison is discussed below:
\par 
\emph{Comparison with Hydra:} As shown in Table \ref{main_table}, while pruning a robust complex model, \emph{Deadwooding} achieves a significantly higher {\sl era} and {\sl eba} as compared to Hydra. Notably, at a 99\% pruning ratio, \emph{Deadwooding} achieves the highest gain. As discussed in Figure \ref{fig:layer-wise}, Hydra’s layer-wise pruning removes 99\% parameters even from the important layers; hence some layers may not have enough retained parameters to model the classification tasks effectively. In contrast, \emph{Deadwooding}’s global pruning retains larger percentages of parameters in important layers while pruning more (up to 99.93\%) in less significant layers, hence achieving consistently high robustness and accuracy.

\par 
\textbf{Approach comparison with most recent SoA baseline Hydra.} 
Finally, the section shed lights on the contrast in performance attained by
\emph{Deadwooding} when compared to Hydra by focusing on analyzing the key conceptual differences between these two approaches.

Firstly, \emph{Deadwooding} prunes the original complex network’s parameters globally while maximizing the projection of input to the decision boundary distances to enforce adversarial robustness. This means that it considers the importance of the parameters from the complete network’s perspective and its attainment of robustness during pruning is not dependent on any specific attack $\oplus$.
In contrast, Hydra performs layer-wise pruning while maximizing adversarial robustness against the PGD attack. This means it needs to remove a relatively higher number of parameters from the important layers, and its robustness is dependent on the specific attack adopted (i.e., PGD). Our analysis shows that the parameters pruned by \emph{Deadwooding} are $10^3$ times smaller, in magnitude, than the retained ones. In contrast those removed by Hydra are just $10$ times smaller in magnitude than the retained ones. 

Finally, Hydra uses natural re-training as the fine-tuning step, and \emph{Deadwooding} uses a modified Knowledge Distillation (with additional enforcement of adversarial robustness), resulting in higher {\sl era}, while achieving similar (marginally higher or lower) {\sl eba}. 
These characteristics allows Deadwooding to extract robust and accurate pruned models, even at a high pruning ratio (99\%).
\vspace{-1.0mm}
\subsection{Delving Deeper into the \emph{Deadwooding}}
This section evaluates the importance of \emph{Deadwooding}’s different components and their characteristics. 
The evaluations are performed on VGG16 trained over the CIFAR-10 dataset. 

\paragraph{\emph{Deadwooding}’s pruning \& fine-tuning components.} We make the following observations:
\begin{wrapfigure}{r}{0.32\linewidth}
\vspace{-12pt}
\begin{minipage}[]{0.99\linewidth}
\centering
\captionsetup{width=.99\linewidth}
\resizebox{.99\columnwidth}{!}{
\begin{tabular}{c | rr rr}
    \toprule
    {Pruning} 
    & \multicolumn{2}{c}{Without $\cL$}
    & \multicolumn{2}{c}{With $\cL$} \\
    \%age & {\sl eba} & \rev{{\sl era}} & {\sl eba} & \rev{{\sl  era}}\\
    \midrule
     95\%& 10&\rev{10} & 82.87 & \rev{59.8}\\
     99\%& 10&\rev{10} & 80.87&\rev{53.1}\\
    \bottomrule
\end{tabular}}
\captionof{table}{Evaluating the significance of $\cL$ during pruning step}
\label{table2}
\end{minipage}
\vspace{-12pt}
\end{wrapfigure}
$\bullet$ \emph{Inclusion of benign accuracy through loss $\cL$.} 
As discussed in Section \ref{sec:Model_pruning}, 
the inclusion of benign accuracy in Equation \eqref{eq:lagrangian_relaxation} enables global pruning by ensuring a connected network with valid input to output layer path. 
Table \ref{table2} clearly shows its importance by evaluating {\sl eba} and {\sl era} on a modified loss which ignores the term $\cL$ from Equation \eqref{eq:lagrangian_relaxation} in the pruning step. 
Note that the resulting pruned models become disconnected, which thus causes the low scores. Additional analysis are shown in Appendix D.1.

\noindent
$\bullet$ \emph{Modified Knowledge distillation fine-tuning.} 
\citet{KD_prune} suggests that applying vanilla KD as fine-tuning step restores the benign accuracy in the pruned model. Additionally, \citet{papernot2016distillation} showed that adversarial KD ($\cL_{\text{soft}}\& \cL_\adv$) makes the neural network models further robust. 
Table \ref{table4} shows that the modified KD fine-tuning (Section \ref{sec:fine_tuning}) achieves pruned models of high {\sl eba} and {\sl era}, whereas vanilla and adversarial KD as fine-tuning achieve high {\sl eba} but lower {\sl era}.
Thus, the proposed modified KD step effectively achieves highly robust and accurate pruned models.
\begin{wrapfigure}{r}{0.5\linewidth}
\vspace{-12pt}
\begin{minipage}[]{0.99\linewidth}
\centering
\captionsetup{width=.96\linewidth}
\resizebox{.99\columnwidth}{!}{
\begin{tabular}{p{2cm} |rr rr rr}
    \toprule
    Fine-tuning
    & \multicolumn{2}{c}{90\%} 
    & \multicolumn{2}{c}{95\%} 
    & \multicolumn{2}{c}{99\%} \\
    
    Approaches& {\sl eba} & \rev{{\sl era}} & {\sl eba} & \rev{{\sl era}}& {\sl eba} & \rev{{\sl era}}\\
    \midrule
    Vanilla KD  &89.05&\rev{13.34} & \textbf{89.18}&\rev{12.98} & \textbf{86.49}&\rev{9.7} \\
    Aversarial KD  
     &\textbf{89.41}&\rev{48.85}&88.39&\rev{48.63}&86.26&\rev{46.05}\\

    \emph{Deadwooding}   &83.56&\rev{\textbf{59.68}} & 82.87&\rev{\textbf{59.8}} & 80.87&\rev{\textbf{53.1}} \\
    \bottomrule
\end{tabular}}
\captionof{table}{Comparison of variations of Knowledge distillation during fine-tuning}
\label{table4}
\end{minipage}

\vspace{-12pt}
\end{wrapfigure}
\noindent

$\bullet$ \emph{Pruned models are robust against stronger adversarial attacks}.
Finally, the paper also evaluates the {\sl era} of pruned models for other attacks $\oplus$: AutoAttack \cite{auto-attack}, FAB \cite{fab-attack} and DeepFool \cite{deepfool-attack}: \emph{Deadwooding} achieves 6.69\%, 13.97\% and 52.06\% higher {\sl era} than Hydra, respectively (see Appendix E.2 for details). 
These results demonstrate \emph{Deadwooding}'s ability in achieving robust pruned models under a variety of settings.


\paragraph{\emph{Deadwooding}’s design choices} 
Finally, this section discusses how \emph{Deadwooding}’s design choices have affected achieving robust and accurate pruned models.


\begin{wrapfigure}{r}{7cm}
  \vspace{-12pt}
  \centering
  \includegraphics[width=0.99\linewidth]{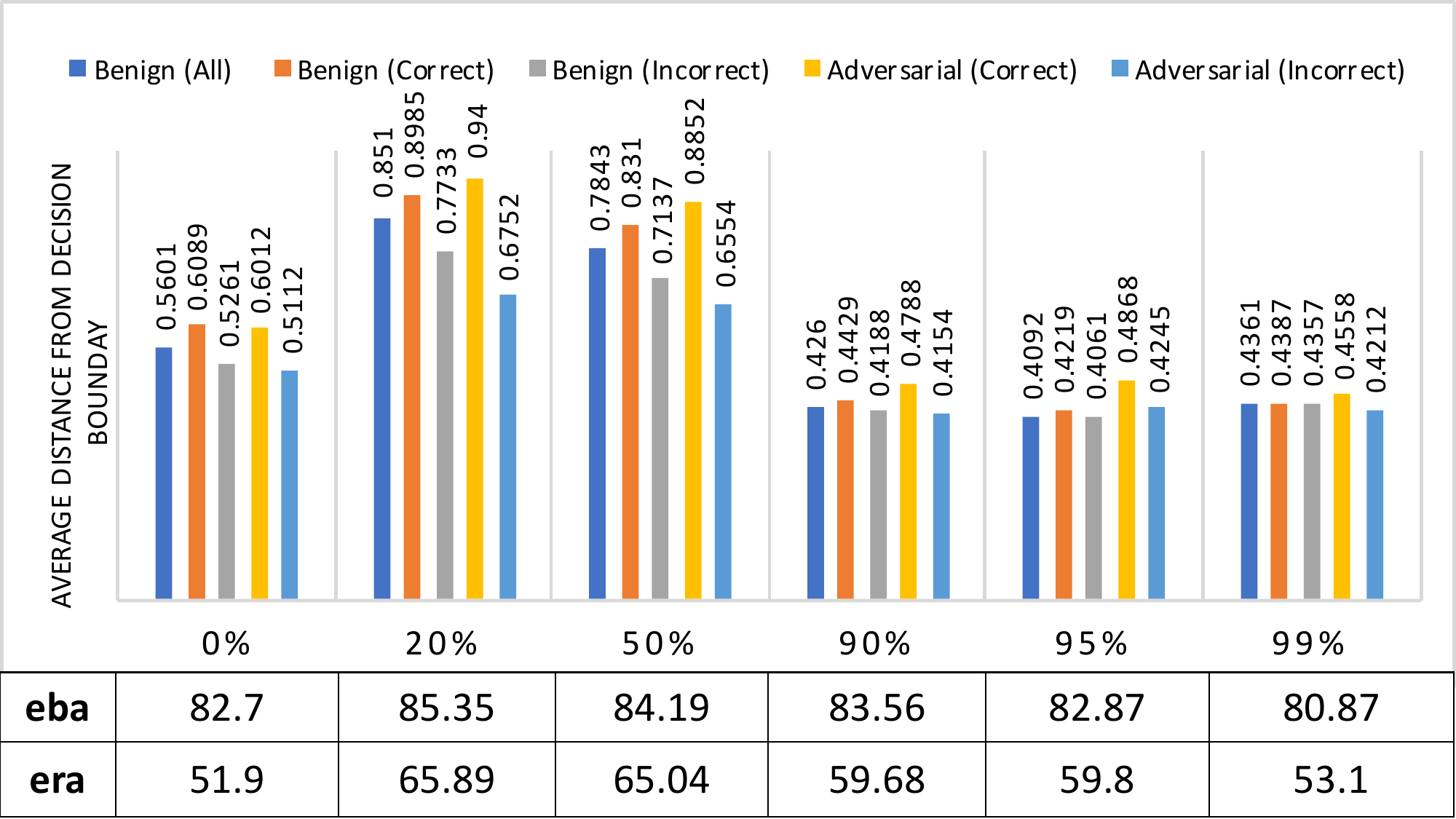}
  \caption{Average distance of input projections to the decision boundaries for various pruning ratios for VGG16 ($M_\theta^r$) trained on CIFAR10 dataset.}
  \label{fig:distance1}
  \vspace{-12pt}
\end{wrapfigure}

$\bullet$ \emph{Using Distance between Projection of Samples and Decision Boundary to impart robustness.}
This section discusses the effectiveness of using distance to decision boundary as a proxy measure to robustness (Equation \eqref{Dist-eq}). 
Figure \ref{fig:distance1} reports the average distance $d,$ $d(X) = 1 - \ell_{\adv}(X)$ (computed as described in Equation \eqref{Dist-eq}) for the test data. 

Higher $d$ values correspond to a higher distance of the samples to the decision boundary. The figure reports the distance to boundary $d$ of the benign and adversarial samples at different pruning amounts for the robust VGG16 model with the CIFAR10 dataset. Note that, on average, correctly classified benign and adversarial samples have higher $d$ than the incorrect ones.
\par
In low pruning ratios (20-50\%), the model goes through the pruning step with distance constraint (Equation \eqref{Dist-eq}) while retaining a large portion of the parameters. In this context, the pruned model is still quite large and the $d$ of the benign and adversarial samples are high. 
\par
However, the model complexity reduces significantly at very high pruning ratios (90-99\%), typically associated with simplified decision boundaries.
Hence, the $d$ values are lower than those obtained in lower pruning ratios.
Notably, {\sl eba} and {\sl era} are relatively high in high-pruning ratios (90-99\%). This is due to the fine-tuning step (Section \ref{sec:fine_tuning}) that re-adjusts the decision boundaries utilizing the retained parameters. Figure \ref{fig:distance1} shows the robustness attained through increasing $d$ in different pruning ratios. The results clearly demonstrate the need of fine-tuning, specifically in high pruning ratios.
\par

\par
$\bullet$ \emph{Lagrangian dual model and Parameter Weights}. 
According to the magnitude-based pruning assumption \citep{Tang2015APB,magnitude_medium}, in a high pruning ratio, the retained parameters should have significantly higher absolute weights, containing substantially more impactful information, compared to the removed parameters.
Notably, in our experiments (not tabulated due to space constraints), after the model pruning step (Section\ref {sec:Model_pruning}) for $90\%, 95\% \text{ and } 99\%$ pruning ratios, the mean absolute weights of the retained parameters are 1100, 1721, 2381 times higher than that of each removed parameter. Additionally, removing the least absolute weight parameters reduces the models' accuracy by at most $0.1\%$, thus, demonstrating their minimal impact. Finally, the Lagrangian dual model adopted achieves highly accurate and robust pruned models, which shows that the retained high mean absolute weight parameters are highly impactful. Thus, the Lagrangian dual model enables effective magnitude-based pruning \citep{Tang2015APB,magnitude_medium}, that is in line with our pruning objective in Equations \eqref{eq:5a}-\eqref{eq:5d}.

\textit{Miscellaneous additional evaluations for Deadwooding are discussed in Appendices E and F.}

\section{Concluding remarks and limitations}

This paper was motivated by the recent observations about the brittleness of pruned models to adversarial attacks. To contrast these observations, the paper presents a novel pruning approach named \emph{Deadwooding} that achieves highly sparse and robust models while maintaining original complex models’ accuracy. 
\emph{Deadwooding} exploits a novel Lagrangian Dual learning method to encourage model sparsity while enforcing robustness. 
The resulting models were shown to outperform state-of-the-art pruning approaches significantly. Notably, \emph{Deadwooding} operates effectively under extremely high pruning-ratios (up to 99\%), and, in contrast with some previous work, it does not leverage any biases or characteristics in the data or domain to attain sparsity and robustness. 

We note that the proposed method is intended to compute pruned models which are robust within a certain bounded region guided by the perturbation budget $\epsilon_\text{max}$, as suggested by \cite{carlini2019evaluating}. However, it does not claim high robustness outside the $L_\infty$ region of radius  $\epsilon_\text{max}$.
While this is a common shortcoming of arguably all practical robust ML models, we believe the is an important avenue of future research.

Despite these avenue of improvement, the authors strongly believe that the presented pruning method will pave the way to the use of robustly pruned ML models in practical safety-critical applications in resource-constrained environments.




\newpage
\bibliographystyle{plainnat}
\bibliography{bibliography}


\appendix

\section{Expermental Details}
\label{Appendix:Experimental-details}
The following section describes the details of the experimental setup:
\par\textbf{Datasets} To demonstrate the effectiveness of our approach \emph{Deadwooding}, the evaluations are carried across three publicly available benchmark dataset: CIFAR-10 with maximum perturbation amount ($\epsilon_\text{max}$) of $8/255$, respectively. For fairness of comparison, the dataset and the corresponding $\epsilon_\text{max}$ are kept consistent with state-of-the-art \cite{sehwag2020hydra,zhang2018systematic}. These datasets can be downloaded and accessed through open-source Torchvision library \cite{10.1145/1873951.1874254}. \par CIFAR-10 is a dataset of 60000 images evenly distributed among 10 mutually exclusive classes, out of which 50000 form the training set and 10000 form the test set. The train and test set reflect the same fair distribution among the 10 classes. 

\par\textbf{Metrics: Benign Accuracy and Robustness.} Empirical Benign Accuracy {\sl eba} and Empirical Robust Accuracy {\sl era} are the two metrics used to measure the performance of a model quantitatively. Benign accuracy {\sl eba} refers to the measure of the accuracy of prediction for the benign (unperturbed) test samples. Several definitions of adversarial robustness exist in literature \cite{robustness_definitions}; however, to maintain consistency, we constrain our evaluations to the definition provided by the SOA baseline \cite{sehwag2020hydra}. Therefore, Empirical Robust Accuracy {\sl era} refers to the measure of the accuracy of models' prediction for the perturbed samples, which are generated using the test set. 
In this study, for generating perturbed samples, we used FGSM-looping during training and the PGD \cite{REN2020346} approach while evaluating (i.e., test) the models to compare with the baseline state-of-the-art approaches. Additionally, we also evaluate the models for stronger attacks like Auto-Attack \citep{auto-attack}, FAB \citep{fab-attack} and Deepfool \citep{deepfool-attack} and the results are presented in Appendix \ref{appendix:strong-attack}.

\textbf{System Configuration} All evaluations are carried over a system with x86\_64 architecture AMD Ryzen 16-core processor, 115 GiB RAM hosting a 16GiB GPU.

\par\textbf{Model Training and parameter configuration}
\emph{Deadwooding} makes use of model training at two stages: pruning and fine-tuning. 

\par During the pruning stage, we leverage Lagrandian dual based optimization (discussed in Section 4) to compute an approximate solution to achieve a combined objective of maximizing robustness and sparsity while maintaining SoA accuracy. The $\lambda$ coefficients for the different components of the loss function are learned within the training process. With each epoch, the model gets more sparse and the amount of sparsity serves as the stopping criterion for this training phase.

\par For fine-tuning, we use modified knowledge distillation which has an additional term for adversarial loss in its loss function (discussed in section 5.2). We used Optuna \cite{akiba2019optuna}, a parameter-tuning tool, to obtain the best values for the coefficients of three loss components: $\alpha, \beta \text{ and} \gamma$ with the objective to maximize {\sl era}. Optuna uses a Bayesian Optimization algorithm called Tree-Structured Parzen estimator to arrive at the optimum set of values. For instance, pruning 99\% parameters of VGG-16 trained for CIFAR-10 dataset used $\alpha = 0.351, \beta = 0.526 \text{ and } \gamma = 0.240$. The resulting pruned model has an $eba = 80.87$ and $era = 53.1$. We ran the fine-tuning step for a maximum of 100 epochs with early-stopping patience of 30 to avoid over-fitting.

\par{\textbf{For fairness of Comparison with SoA}} The results for \citet{sehwag2020hydra} (Hydra) are generated using codes available at public git repository \citet{Hydra_git}, respectively. The original models used for evaluations are taken from the checkpoints available at the public git repository \citet{Hydra_git} provided by \citet{sehwag2020hydra}.

\section{Background Discussion}
This section discusses the details of various terms used in this work. Section \ref{model_pruning} describes model pruning and the most prevalent ways to achieve it \cite{han2015deep,zhang2018systematic,sehwag2019compact,sehwag2020hydra}. Section \ref{adversarial} described adversarial attacks and how adversarial training is used to achieve robustness against such attacks. \emph{Deadwooding} uses a modified form of knowledge distillation during the fine-tuning stage; thus, Section \ref{Knowledge_distillation} is used to discuss the details of vanilla knowledge distillation.

\label{Appendix:Background-discussion}
\subsection{Model Pruning}
\label{model_pruning}
Initial efforts to optimize the inferences made by the machine learning models led to a considerable increase in their size in terms of the number of parameters, which require enormous amount of disk space. For instance, the number of parameters required for VGG-16 trained on CIFAR-10 are 138M.
Parameter pruning techniques tend to achieve the effect of a compressed model by making the model parameters sparse. The conventional method of parameter pruning is to remove the connections (i.e., parameters or neurons) which contribute the least towards the model's inference \cite{sehwag2019compact,zhang2018systematic}. In literature, the importance of the model parameters is measured in terms of the magnitude of their value\citep{Tang2015APB,magnitude_medium}. However, at high pruning amounts, such a removal leads to a loss in accuracy. The retained model parameters are then fine-tuned (i.e., re-trained) to re-gain the lost performance \cite{han2015learning}. 

\textbf{Global vs Layer-wise Pruning:} In literature \cite{han2015deep,zhang2018systematic,sehwag2019compact,sehwag2020hydra}, pruning has been done either \emph{layer-wise} or \emph{globally}. \emph{Layer-wise pruning} checks for the least contributing connections belonging to a single layer at a time, whereas \emph{global pruning} considers all the model connections simultaneously for removal of least significant parameters. It has been shown recently that for the same amount of pruning, the model pruned using global pruning exhibit better accuracy of prediction than a model pruned using layer-wise pruning \citep{aghli2021combining}. Since global pruning does not restrict the fraction of parameters pruned or retained for each layer, the network pruning solely based on the magnitude of weights may result in disconnected networks for some instances. These disconnected networks cannot regain their performance, even after fine-tuning. To avoid such circumstances, state-of-the-art approaches \citet{sehwag2020hydra,sehwag2019compact,zhang2018systematic} followed layer-wise pruning, which removes a fixed amount of parameters from each layer so that each layer retains some of its parameters. We, however, prune the networks globally, resulting in different pruning amounts at each layer. Moreover, while globally pruning the network, we ensure to retain the model's accuracy, thus, resulting in at least one valid path from the input layer to the output.

\subsection{Adversarial attacks and robustness}
\label{adversarial}
\citet{goodfellow2015explaining} demonstrated that small perturbation to input samples can result in incorrect model inference with high confidence. A perturbed sample is generated by adding a well-crafted noise to the sample so that an oracle (or human observer) perceives it as the original. The perturbed samples which get misclassified by the target model are considered adversarial samples \cite{goodfellow2015explaining}. Several \emph{adversarial attacks} like FGSM \citep{goodfellow2015explaining}, PGD \citep{madry2018towards}, DeepFool \citep{deepfool}, etc. have been proposed in the literature to compute worst-case adversarial samples with minimum perturbations. \emph{A model which can correctly classify these perturbed samples is said to possess adversarial robustness.}

\begin{wrapfigure}{r}{5 cm}
\includegraphics[width=5cm]{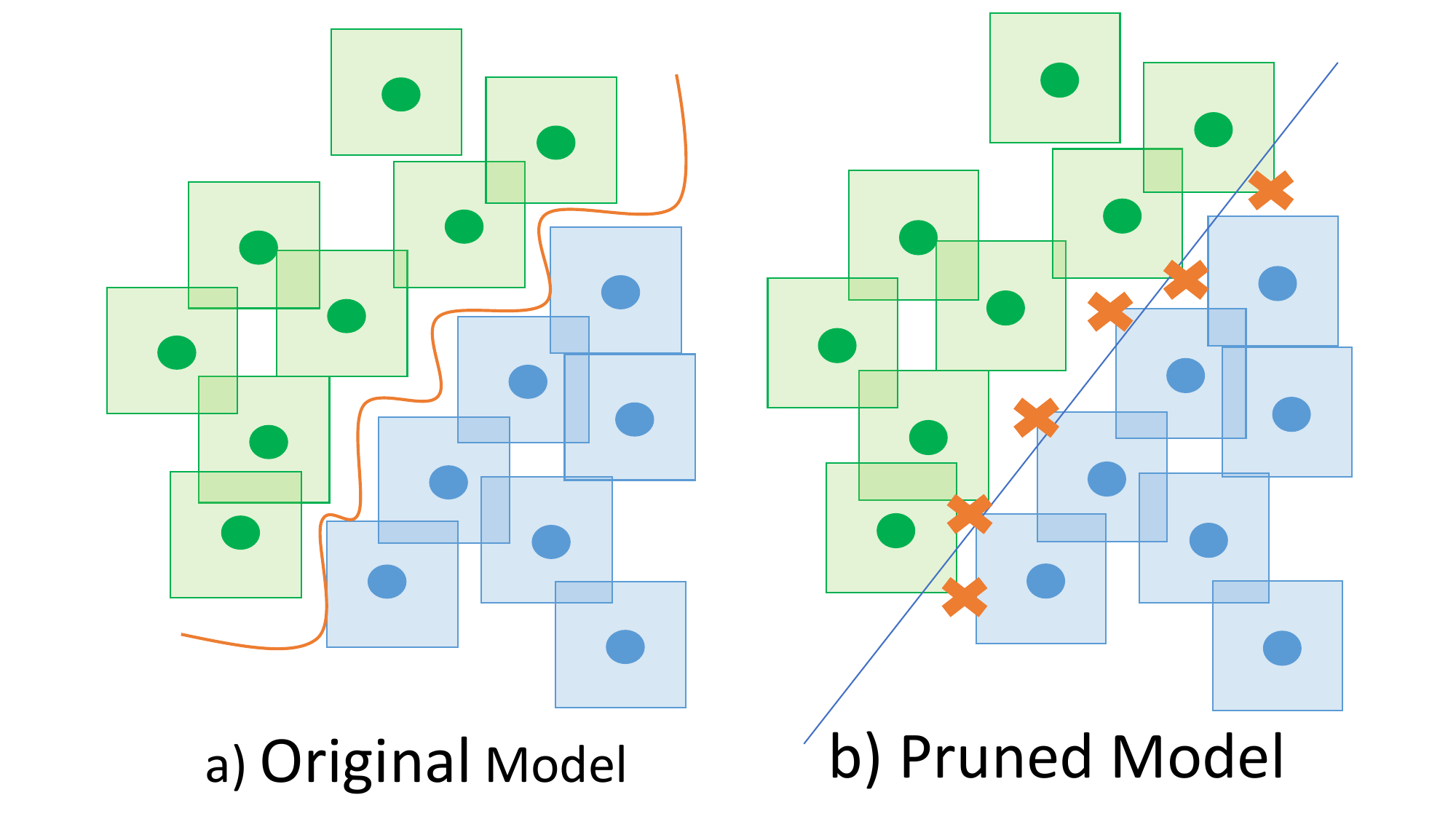}
  \vskip -1ex
  \caption{Pruned Models are more vulnerable to adversarial perturbaion}
  \vskip -4ex
  \label{fig:vulnerability}
\end{wrapfigure}

\textbf{Pruning reduces robustness} Figure \ref{fig:vulnerability} shows a depiction of the projection of samples belonging to two classes, each surrounded by $L_\infty$ region of a fixed radius (known as perturbation budget), where a corresponding perturbed sample can lie. Figure \ref{fig:vulnerability} (a) represents a complex decision boundary learned by the original complex model. Lesser number of parameters limits the ability of a pruned model to learn this complex function and the decision boundaries thus learned are quite simple. Figure \ref{fig:vulnerability} (b) shows that for a pruned model, when a perturbation is introduced, the samples closer to the decision boundary are more prone to incorrect classification because of the simplified decision boundary.

\textbf{Adversarial training} Adversarial training aims at enabling the machine learning models to be adaptive to small perturbations to benign samples. Adversarial examples are generated and used when training the model. Intuitively, if the model sees adversarial examples during training, its performance at prediction time will be better for adversarial examples generated in the same way.
Different attack generation methods, such as the fast gradient sign method (FGSM), projected gradient descent (PGD), etc. \cite{wong2020fast}, can be employed to generate adversarial examples during training. Adversarial training aims to minimize the loss $\cL_\adv$ as described in Section 3.

\subsection{Knowledge Distillation (KD)}
\label{Knowledge_distillation}
Vanilla KD aims to transfer the knowledge learned by one model, called as teacher model, to another model, named the student model. The student model is trained to achieve a combined objective of high accuracy and learning the teacher model's inference space distribution. Minimizing the loss functions $\cL_{KD}$ is the key behind the training of the student model.
\begin{equation}
\begin{aligned}
    \cL_{KD} &= \alpha * \cL + (1-\alpha) * \cL_{\text{soft}} \\
    \cL &= \cL((\theta_s,x,y,\hat{y})) \\
    \cL_{\text{soft}} &= \cL(f_T(\theta_s,x),f_T(\theta_t,x))
\end{aligned}
\label{Background_KD_LOSS}
\end{equation}
Here $\theta_t$ and $\theta_s$ represent the parameters of the teacher model and the student model, respectively. $x$ is an input sample, $y$ is the designated label, and $\hat{y}$ is the prediction made by the student model. Thus, $\cL$ aims to achieve accuracy by minimizing the difference between the true labels and the predictions made by the model.
\par
Softmax of model output is computed using the formula $f_T(\theta,x) = \frac{e^{M_\theta(x)/T}}{\Sigma_{k=1}^n e^{M_\theta(x)/T}}$, by substituting T = 1. Using higher values of T, allows for smoother probability distribution in output which reveals the relation among all the classes. In Equation \ref{Background_KD_LOSS}, $f_T(\theta_t,x)$ and $f_T(\theta_s,x)$ correspond to the logit values generated using softmax at temperature $T$ by the teacher and student models \cite{hinton2015distilling}. $\cL_{\text{soft}}$ enables the student model to learn response based knowledge of the teacher model, provides better generalization, and helps to optimize the student model leveraging a small amount of training data \cite{gou2021knowledge}. Several pruning approaches leveraged KD to fine-tune the retained pruned network \cite{aghli2021combining,prakosa2021improving}.
Notably, a previous work \cite{papernot2016distillation}, used knowledge distillation to enhance the robustness of a model having the same architecture as that of the teacher model.

\section{Related Works: Robust Pruning}

Initial efforts that used model pruning as a model compression mechanism targeted maintaining accuracy while pruning the redundant parameters. \citet{han2015deep} used the least-weight-magnitude-based global pruning mechanism and achieved a maximum pruning amount of 34\%. \citet{Li2017PruningFF} focused on pruning conv-nets by removing filters instead of independent parameters for convolutional layers pruning a maximum of 38.1\% parameters. \citet{zhang2018systematic} defined the pruning problem as a non-convex optimization problem, solved it using the ADMM approach while adopting layer-wise pruning, and demonstrated a maximum pruning of 71\%. In contrast to these approaches, \emph{Deadwooding} targets to achieve highly pruned networks (90\% - 99\%) while maintaining accuracy as well as amplifying robustness.

Recent studies \cite{ramanujan2020s,frankle2018lottery} established the existence of a randomly initialized sub-network in a larger model matching the benign accuracy of a pre-trained network of similar complexity. Additionally, \citet{ramanujan2020s} demonstrated that highly robust sub-networks exist within complex networks. SoA approaches \citet{sehwag2019compact,sehwag2020hydra} aimed to compute robust pruned networks using a layer-wise pruning approach to remove an equal percentage of parameters from each layer, achieving a maximum of 90\% and 99\% pruning, respectively. However, recent works have demonstrated that for the same pruning amount, pruned models obtained using global pruning have better prediction accuracy than the models pruned layer-wise\citep{aghli2021combining}. 
\par 
Some other works focused on different aspects of compression, such as quantization \cite{polino2018model}, Low-Rank Factorization \cite{swaminathan2020sparse}, distillation \cite{sau2016deep}, etc. In-time sparsification is one of the emerging approaches which intend to train an already sparse network while constraining the number of parameters, resulting in a sparse network with a different configuration than the original model \cite{in-time}. However, this paper presents a novel robust pruning-based model compression mechanism since it has the advantage of using already trained complex off-the-shelf models and has been widely used in literature \cite{Han2021PreTrainedMP}. Thus, our SoA constitutes only model pruning approaches.


\section{Importance of various components of \emph{Deadwooding}}
This section aims to draw attention to the significance of some components of \emph{Deadwooding} which results in a substantial improvement over the state-of-the-art. Section \ref{Appendix:global-pruning} discusses the role of benign accuracy during sparsification and how it helps to achieve global pruning. Section \ref{appendix:layer-importance} demonstrates that \emph{Deadwooding} captures the importance of different layers of the model towards model inference. Finally, section \ref{appendix:fgsm-looping} is a continuation of section 5.2 and explains the benefits of using FGSM-looping (a variation of FGSM) as an adversarial loss component during fine-tuning.

\subsection{Achieving Global Pruning through Benign Accuracy (Continuation of Section 6.2)} \label{Appendix:global-pruning}

We now show evaluations that demonstrate the importance of using benign accuracy in the sparsification training step (Equation (4) in the main paper) and how it enables global pruning while maintaining network connectivity. We also discuss the impact of using global pruning instead of restricted layer-wise pruning on the model's performance. The results presented here are computed for a VGG-16 network trained for the CIFAR-10 dataset when it is pruned by 99\% using various approaches.
\par

\begin{figure}
\centering
\begin{subfigure}{.5\textwidth}
  \captionsetup{width=.95\linewidth}
  \centering
  \includegraphics[width=0.95\linewidth]{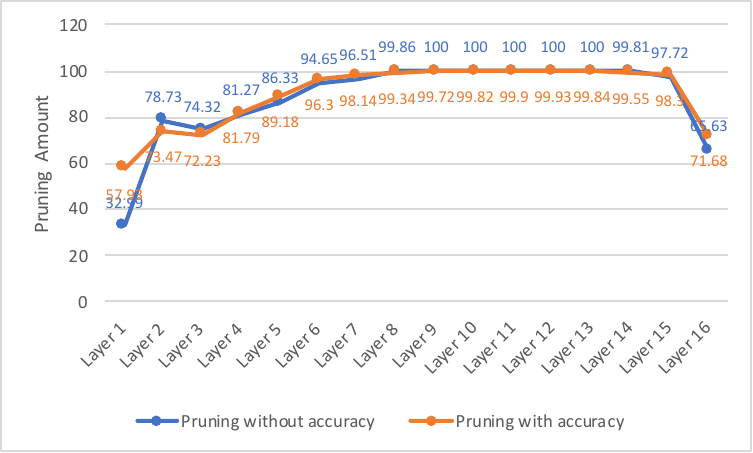}
  \caption{Layer wise pruning amounts with and without considering accuracy while pruning for VGG16 with CIFAR10 dataset at 99\% pruning amount}
  \label{fig:importance_acc}
\end{subfigure}%
\begin{subfigure}{.5\textwidth}
  \captionsetup{width=.95\linewidth}
  \centering
  \includegraphics[width=0.95\linewidth]{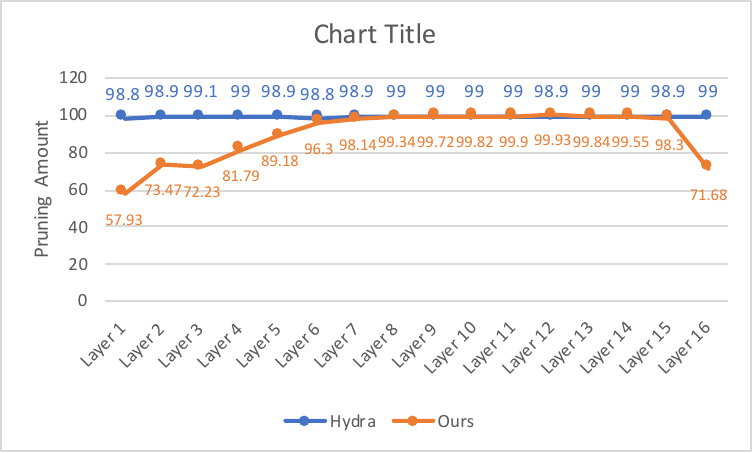}
  \caption{Layer wise pruning amounts for \emph{Deadwooding} and Hydra for pruning for VGG16 with CIFAR10 dataset at 99\% pruning amount}
  \label{fig:pruning_hydra_vs_ours}
\end{subfigure}
\caption{Importance of using benign accuracy as a loss component during sparsification training}
\end{figure}

Figure \ref{fig:importance_acc} shows the amount of pruning incurred at each layer using two variations of \emph{Deadwooding} corresponding to the presence and absence of the term \emph{benign accuracy} ($\cL$) in the loss function used during sparsification training step. According to this comparison, the pruned model obtained without using $\cL$ is disconnected and possesses no remaining connections (i.e., parameters) in layers $9-13$. Also, the accuracy of the model, thus obtained, is just $10\%$ (as shown in Section 6, Table 4). In contrast, our presented approach (i.e., including $\cL$ in sparsification loss) retains parameters in all layers contributing to active input-to-output path, resulting in a highly accurate connected sub-network.  
\par 
Furthermore, Figure \ref{fig:pruning_hydra_vs_ours} shows the comparison of the pruning percentage at each layer of the pruned model obtained using Deadwooding vs. Hydra. Hydra performs layer-wise pruning; hence the same percentage of parameters are removed from each layer. In contrast, Deadwooding performs global pruning; as a result, each layer of the pruned model has a different percentage of parameters removed. Since not necessarily every layer is similarly contributing to the model’s inference, Deadwooding’s global pruning effectively captures the disproportionate layer importance and achieves pruned models which exhibit high benign accuracy and robustness. Notably, for VGG-16, the initial and the final layers undergo a lesser pruning amount than the middle layers.

\subsection{Pruning parameters accoring to the importance of a layers} \label{appendix:layer-importance}

\begin{wrapfigure}{r}{0.6\linewidth}
\vspace{-12pt}
  \centering
  \includegraphics[width=0.95\linewidth]{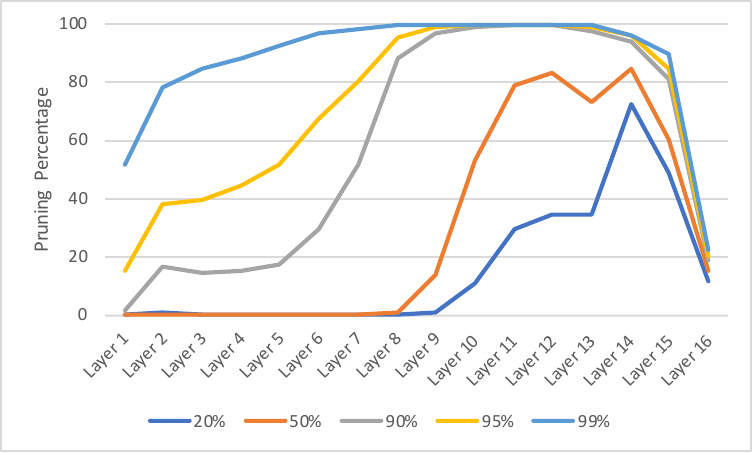}
  
  \caption{Layer wise pruning amounts for multiple pruning amounts for VGG16 with CIFAR10 dataset}
  \label{fig:pruning_multiple}
  \vspace{-12pt}
\end{wrapfigure}

Figure \ref{fig:pruning_hydra_vs_ours} exemplifies that initial layers $1-6$ and the last layer are relatively more important in the 99\% pruned VGG16 model on the CIFAR10 dataset. To further investigate, we compare the amount of pruning in each layer for five different pruning ratios. Following observations can be inferred from the results shown in Figure \ref{fig:pruning_multiple}:
\begin{enumerate}

\item For smaller pruning amounts (20-50\%), the layers 13-15 get pruned in a significantly higher percentage than others. This is because they are fully connected linear layers and have more redundancy than the initial convolution layers that extract meaningful features for effective classification.

\item As the required pruning increases, the pruning amount of the convolution layers also increases with the depth of the layer. This is because the initial convolution layers extract low-level features required in the following layers for high-level features extraction. Hence, initial layers are pruned lesser than the inner convolution layers. 

\item The last layer, which decides the label of the input, gets pruned the least among all the linear layers.
\end{enumerate} 

The state-of-the-art approaches \citep{sehwag2020hydra,sehwag2019compact,zhang2018systematic} which use layer-wise pruning, prune all the layers by an equal amount without considering their overall importance. This results in a loss of information learned by the original complex model, leading to compromised model performance. \emph{Deadwooding}, however, incurs different pruning amounts per layer, and the above analysis shows that these pruning amounts depend on the importance of each layer towards the model's inference. The pruned models thus produced by SoA approaches have lower {\sl eba} and {\sl era}.

\subsection{Two-fold benefit of using FGSM-Looping (Continuation of Section 5.2)} \label{appendix:fgsm-looping}

\begin{wrapfigure}[10]{r}{6.2cm}
\begin{flushright}
\begin{minipage}{6.0cm}
\vspace{-16pt}
\begin{algorithm}[H]
{\small
  \caption{$\oplus$ operator}
  \label{alg:fgsm-looping}
  \setcounter{AlgoLine}{0}
  \SetKwInOut{Input}{input}

  \Input{ $\theta$: Model parameters 
  $\diamond~(X, Y)$: training sample
  $\diamond~E$: Set of pertubations $\leq \epsilon_{\text{max}}$
  $\diamond~t$: epoch}
    $\epsilon = E[t (mod |E|)]$\\
    Perturbed sample, $X_i' = {\text FGSM}(\theta,X_i,Y_i,\epsilon)$\\
    return $X_i'$
  }
\end{algorithm}

\end{minipage}
\end{flushright}
\end{wrapfigure}

We used two techniques to generate adversarial examples in our work, namely, FGSM-looping (a variation of the Fast Gradient Sign Method (FGSM)) during fine-tuning and Projected Gradient Descent (PGD) during the evaluation of models. FGSM-looping is a variation of FGSM attack, where we iterate over a set of values of perturbation to be used in different epochs. Algorithm \ref{alg:fgsm-looping} describes the procedure for FGSM-looping.

FGSM is a single-step attack done on a sample, which generates a perturbation at a distance $\epsilon_{\text{max}}$ in the direction of the gradient for that sample. The perturbed samples generated by FGSM are restricted to $\pm\epsilon$ perturbation for each feature which bounds the perturbed samples to be on the corners of an $L_\infty$ ball of radius $\epsilon_\text{max}$ around the sample.  Although the direction of the gradient for a sample may change in every training epoch, which helps FGSM to generate different perturbations for the same benign sample, but for a $d$-dimensional feature space, the size of this search space for FGSM is $2^d$. Many recent works discouraged the use of FGSM for adversarial training because of its restricted search space\citep{madry2018towards,no_fgsm}.

PGD, however, is a multi-step attack, which generates a small amount of perturbation in one iteration and computes the gradient again to change the direction of perturbation in each step while keeping the perturbed sample in $L_\infty$ ball of radius $\epsilon_{\text{max}}$. Thus, PGD searches for the perturbed sample in whole of the $L_\infty$ space. As a result, PGD requires more computation time than FGSM to generate one perturbed sample.

\citet{madry2018towards} suggests that for small adversarial perturbations in $L_\infty$ norm, FGSM can produce adversarial samples which are similar to the ones produced by PGD. Motivated by that, FGSM-looping helps achieve the same effect as PGD by generating adversarial examples for multiple values of $\epsilon$ ranging from $\sim0$ to $\epsilon_\text{max}$; thus, increasing the search space of FGSM. Therefore, with FGSM-looping, we get an additional benefit of varying magnitude in addition to varying the direction of the perturbation in every epoch. As a result, we are able to achieve similar results as compared to PGD (step size = 10) adversarial training, but in three times lesser computation time (for the current system configuration).

\section{Miscellaneous Results}
This section discusses some miscellaneous results which support the results presented in the main paper. 
Section \ref{appendix:strong-attack} shows the {\sl era} evaluations of the pruned models against stronger adversarial attacks. 

\subsection{Testing with Stronger attacks for same perturbation budget (Continuation of Section 6.2)}
\label{appendix:strong-attack}
This section aims to demonstrate the ability of the pruned model to defend against stronger attacks.

Table \ref{strong_attack} compares the results for {\sl era} for stronger attacks on a 99\% pruned VGG-16 model trained for CIFAR-10 obtained using Hydra and Deadwooding. The evaluations demonstrate that \emph{Deadwooding} outperforms Hydra by 6.69\%, 13.97\% and 52.06\% {\sl era} for attacks  AutoAttack\citep{auto-attack}, FAB\citep{fab-attack} and Deepfool \citep{deepfool-attack}, respectively. Notably, the perturbed samples generated by DeepFool\cite{deepfool-attack} are not bounded by a perturbation budget ($\epsilon_\text{max}$), and Deadwooding outperforms Hydra in this case as well. 

The stronger versions of PGD attacks tend to find adversarial examples in a fine-grained manner by increasing the number of attack steps $numSteps$ while keeping the maximum perturbation $\epsilon_{max}$ same. The step size $\alpha$ is computed using $\alpha = 2.5 * \epsilon_{max}/numSteps$ which is in line with the state-of-the-art \cite{madry2019deep}.
Table \ref{strong_PGD_attack} shows results for {\sl era} for five pruning amounts, which are 99\%, 95\%, 90\%, 50\% and 20\%, for VGG16 on CIFAR-10 dataset. We show the results for four variations of the $numSteps$, which are 10, 50, and 100.  The evaluation shows that even for stronger PGD attacks, we achieve a similar amount of {\sl era} with an average standard deviation of $0.0062$ for all the pruning amounts.

\begin{figure}[!t]
\begin{minipage}{0.5\linewidth}
\centering
\captionsetup{width=.95\linewidth}
\resizebox{.95\columnwidth}{!}{
\begin{tabular}{c|c|c|c}
    \hline
    Attack & Auto-Attack & FAB & DeepFool\\
    \hline
    Hydra& 56.66&70.96&27.45\\
    Deadwooding& 60.45&80.87&41.74\\
    \hline
\end{tabular}}
\captionof{table}{Comparison of {\sl era} for different attacks on 99\% pruned VGG-16 model trained for CIFAR-10}
\label{strong_attack}
\end{minipage}
\begin{minipage}{0.5\linewidth}
\centering
\captionsetup{width=.95\linewidth}
\resizebox{.95\columnwidth}{!}{
\begin{tabular}{c|c|c|c|c|c|c}
    \hline
    Step Size & 10 & 50 & 100 & Average& Std dev\\
    Pruning\%&&&&&&\\
    \hline
    20\%&65.89&65.95&65.82&65.88&	0.0028\\
    50\%&65.04&64.96&64.92&64.97&	0.0025\\
    90\%&59.68&59.61&59.64&59.64&	0.0032\\
    95\%&59.8&59.62&59.92&59.78&	0.0149\\
    99\%&53.1&52.92&52.88&52.99&	0.0097\\
    \hline
\end{tabular}}
\captionof{table}{Comparison of {\sl era} for PGD attack with different step sizes for VGG16 and dataset CIFAR-10 at 5 pruning amounts.}
\label{strong_PGD_attack}
\end{minipage}

\end{figure}

\section{Comparison with Other Pruning Techniques}

This section demonstrates the effectiveness for pruning mechanism used in \emph{Deadwooding} as compared to some other pruning paradigms presented in literature: training from scratch (Section \ref{appendix:from-scratch}), multi-step pruning (Section \ref{appendix:multi-step}) and least weight magnitude based pruning (Section \ref{Appendix:lwm-pruning}).

\subsection{Comparison with training from scratch (Continuation of Section 7)} \label{appendix:from-scratch}
\par
Leveraging \emph{Deadwooding}, training a smaller network from scratch can be considered a more straightforward approach (than pruning) to achieve compressed models. However, several prior works \cite{li2016pruning,han2015learning} showed that training a pruned model from scratch does not result in an efficient model and that retaining parameters from the initial training phase results in a better solution. 
\par 
To evaluate that, we applied \emph{Deadwooding} on a randomly initialized VGG16 model as the base model instead of a pre-trained one. The evaluation was performed on the CIFAR-10 dataset at 99\% pruning amount. According to the evaluation, the resulting compressed model fails to achieve acceptable accuracy (only 10\% {\sl eba}); hence demonstrates that training the compressed model from scratch is not effective, which is in line with the state-of-the-art \cite{sehwag2020hydra,zhu2017prune}.

\subsection{Comparison with Multi-step pruning (Continuation of Section 7)}\label{appendix:multi-step}
This section demonstrates that \emph{Deadwooding} results in pruned models exhibiting similar {\sl eba} and {\sl era} for single-step and multi-step pruning (also known as Iterative Pruning \cite{molchanov2017pruning}). In multi-step pruning, each iteration constitutes two steps: Pruning and Fine-tuning. We start with a low pruning amount and keep increasing the pruning amount in each iteration until the desired pruned amount (in the compressed model) is achieved. We compared the evaluation metrics of the 90\% pruned VGG16 model on the CIFAR-10 dataset using the presented single-step pruning approach and the multi-step version of our approach. We observe that the evaluation metrics for both approaches are comparable, with a variation of 0.3 percent points in the {\sl eba} and {\sl era}. Thus, we conclude that the multi-step pruning and single-step pruning result in similar pruned models, which is in line with state-of-the-art \cite{sehwag2020hydra}.

\subsection{Comparison with Least Weight Magnitude based pruning}
\label{Appendix:lwm-pruning}
The Least Weight Magnitude-based pruning is a gold-standard pruning step technique used by several prior studies \cite{sehwag2020hydra,Zhang_2018,sehwag2019compact}. To compare least weight magnitude-based pruning with the Deadwooding’s model pruning step, we performed least weight magnitude-based pruning (instead of \emph{Deadwooding's} sparsification step) followed by our modified adversarial knowledge distillation fine-tuning step on the VGG16 network (99\% pruning ratio and on the CIFAR10 dataset). The pruned model was disconnected and achieved only a 10\% {\sl eba}, evidencing that our presented global pruning step is more effective than the LWM-based global pruning approach.

\end{document}